\documentclass[times,review,10pt]{elsarticle}
\usepackage[top=4.3cm, right=4.8cm, bottom=4.3cm, left=4.8cm]{geometry}

\usepackage{amssymb}
\usepackage{amsmath}
\usepackage{booktabs} 
\usepackage{multirow}
\usepackage{multicol}
\usepackage{adjustbox}
\usepackage{color}
\newcommand{\Hao}[1]{{\color{blue}{#1}}}

\journal{Nuclear Physics B}

\begin{document}

\begin{frontmatter}



\title{Faithful Extreme Image Rescaling with SVD-Initialized Free Channel Mixing and Semantic Priors} 

\author[label1]{Hao Wei\corref{cor1}}
\cortext[cor1]{corresponding author}
\ead{cshaowei2021@gmail.com}
\author[label1]{Yanhui Zhou}
\author[label1]{Chenyang Ge}
\affiliation[label1]{organization={Xi'an Jiaotong University},
             addressline={No.28, West Xianning Road}, 
             city={Xi'an},
             postcode={710049},
             state={Shaanxi},
             country={China}}
\author[label2]{Saeed Anwar}
\author[label2]{Ajmal Mian}
 \affiliation[label2]{organization={The University of Western Australia},
             city={Perth, Crawley},
             postcode={6009},
             state={WA},
             country={Australia}}



\begin{abstract}
\Hao{Extreme image rescaling is essential for bandwidth-constrained visual sensing and robotic perception. However, aggressive downscaling by factors of $16\times$ or higher introduces severe information bottlenecks that degrade structural coherence, texture fidelity, and semantic consistency, thereby limiting downstream pattern recognition tasks. Because low- to high-resolution mapping is severely ill-posed at such extreme scales, most existing rescaling methods struggle to preserve semantically consistent structures and produce realistic details.}
To alleviate the above problems, we propose FaithEIR, a diffusion-based framework for extreme image rescaling. Inspired by singular value decomposition, we develop SVD-initialized free channel mixing that enables adaptively downscaling and upscaling in the latent space. To compensate for information loss due to quantization, we propose an adaptive detail prior, a high-frequency dictionary that captures the empirical average of commonly occurring structures in the training data. Finally, we design a lightweight pixel semantic embedder to provide semantic conditioning for the pretrained diffusion model. We present extensive experimental results demonstrating that our FaithEIR consistently outperforms state-of-the-art methods, achieving superior reconstruction fidelity and perceptual quality. Our code, model weights, and detailed results are released at https://github.com/cshw2021/FaithEIR.
\end{abstract}


\begin{keyword}
Extreme image rescaling \sep diffusion model \sep free channel mixing \sep adaptive detail prior \sep semantic conditioning 
\end{keyword}

\end{frontmatter}



\section{Introduction}\label{introduction}
The proliferation of ultra-high-definition imaging devices has led to a rapid increase in such images, posing significant challenges for data storage, visualization, and transmission.  This trend has driven growing demand for image rescaling techniques that downscale high-resolution (HR) images to visually valid low-resolution (LR) images and then upscale them to recover the original HR images. The LR images play a critical role in reducing storage and bandwidth requirements.

\begin{figure}[!t]
\footnotesize
\centering
    \begin{tabular}{ccc}
            \multicolumn{3}{c}{
            \includegraphics[width=0.98\linewidth,height=0.6\linewidth]{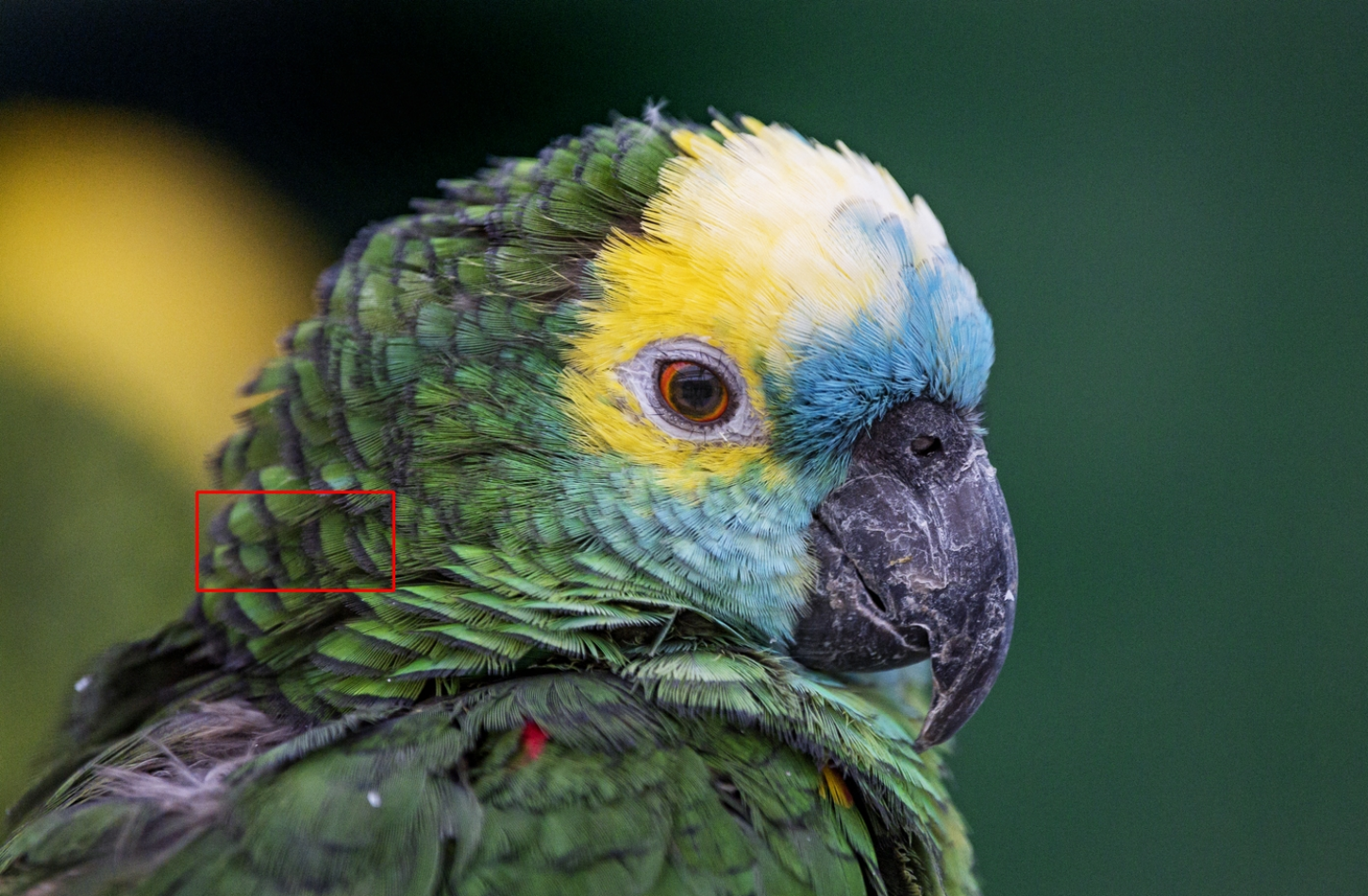}} \\
            \makebox[0.98\linewidth]{HR image} \\
            \includegraphics[width=0.32\linewidth,height=0.21\linewidth]{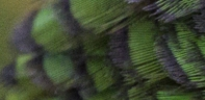}
            \includegraphics[width=0.32\linewidth,height=0.21\linewidth]{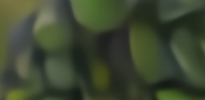}
            \includegraphics[width=0.32\linewidth,height=0.21\linewidth]{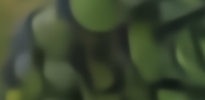}
              \\
            \makebox[0.32\linewidth]{(a) HR patch}
            \makebox[0.32\linewidth]{(b) IRN} 
            \makebox[0.32\linewidth]{(c) T-IRN} \\
            \includegraphics[width=0.32\linewidth,height=0.21\linewidth]{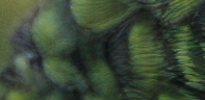}
            \includegraphics[width=0.32\linewidth,height=0.21\linewidth]{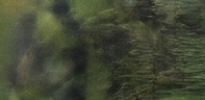}
            \includegraphics[width=0.32\linewidth,height=0.21\linewidth]{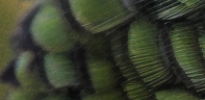} \\
            \makebox[0.32\linewidth]{(d) VQIR}
            \makebox[0.32\linewidth]{(e) TADM}
            \makebox[0.32\linewidth]{(f) FaithEIR} \\
    \end{tabular}
\caption{Comparison with state-of-the-art rescaling methods. (b)-(f) present reconstruction results with a scale factor of 16$\times$. Compared with competing methods, the proposed FaithEIR produces more realistic and faithful details.}
\label{vis_com_intro}
\end{figure}

Early image rescaling methods typically formulate the task as super-resolution, where HR images are first downscaled using predefined kernels (e.g., bicubic interpolation) and then upscaled using super-resolution networks~\citep{EDSR,SRReview_PR2025, RCAN, SSIR_PR2024}. However, such approaches decouple the downscaling and upscaling processes, overlooking their potential for joint optimization and consequently failing to adequately address the ill-posed nature of image upscaling.

To alleviate this issue, recent image rescaling methods have been developed. These can be broadly categorized into two groups: (i) approaches that learn downscaling operators optimized for subsequent upscaling~\citep{TAD_TAU,CNN_CR,CAR}, and (ii) methods based on invertible neural networks~\citep{IRN,HCFlow,HfFlow,T-IRN}. The former typically employ separate neural networks for downscaling and upscaling, whereas the latter formulate image rescaling as a reversible process in which the forward (downscaling) and backward (upscaling) transformations are implemented via invertible neural networks~\citep{Glow}. However, when applied to extreme rescaling factors (e.g., $16\times$ and $32\times$), these methods remain dominated by pixel-wise optimization objectives, often leading to overly smoothed reconstructions that preserve only coarse structural information, as illustrated in Fig.~\ref{vis_com_intro}(b) and (c).

Recent advances in generative adversarial networks~\citep{StyleGAN,VQGAN} and diffusion models~\citep{SD-Turbo,DiffEIC,RDEIC,AddSR_PR2026} have demonstrated that strong priors encoded in pretrained generative models can provide rich and diverse representations for extreme image upscaling~\citep{GRAIN}. For instance, VQIR~\citep{VQIR} exploits visual embeddings learned by a pretrained VQGAN, while TADM~\citep{TADM} leverages the powerful generative capabilities of large-scale text-to-image diffusion models. However, these methods often suffer from undesirable hallucinations, leading to distorted structures and semantic inconsistencies in the reconstructed images (see Fig.\ref{vis_com_intro}(d) and (e)). \textit{These observations highlight the need for extreme rescaling frameworks that can effectively harness generative priors while preserving structural fidelity and suppressing hallucinations in HR reconstructions.}

In this paper, we propose an effective diffusion-based rescaling framework, termed FaithEIR, which performs image rescaling in the latent space. Inspired by singular value decomposition, we develop an SVD-initialized free channel mixing (SVD-FCM) module for adaptive downscaling and upscaling that outperforms the widely used Haar transform used in previous rescaling methods. Instead of explicitly modeling lost information using a normal distribution or a learnable Gaussian Mixture Model, we propose an adaptive detail prior (ADP), a simple yet effective high-frequency dictionary learned from training data to capture typical structural patterns. In addition, to fully exploit the generative capabilities of pretrained diffusion models, we design a lightweight pixel semantic embedder (PSE) that extracts semantic cues from the LR image. These cues serve as effective conditioning signals for the diffusion model, facilitating more faithful feature restoration and improved reconstruction quality. As shown in Fig.~\ref{vis_com_intro}(f), the proposed FaithEIR generates more faithful and high-fidelity reconstructions that are better aligned with the original HR image.

We summarize our main contributions as follows:
\begin{itemize}
    \item We propose a diffusion-based extreme image rescaling method that performs rescaling in the latent space and demonstrate that it quantitatively and qualitatively outperforms state-of-the-art approaches.
    \item We introduce an SVD-initialized free channel mixing module that combines an orthogonal starting point with unconstrained training, achieving adaptive downscaling and upscaling of latent features.
    \item An adaptive detail prior is designed to compensate for lost information by learning typical average structural patterns from the training data.
    \item To provide effective semantic guidance for the diffusion model, we design a pixel semantic embedder that extracts semantic information from the LR image and enforces semantic consistency with the corresponding HR image via a semantic alignment loss.
\end{itemize}
\section{Related work}
\subsection{Conventional Image Rescaling}
Rescaling aims to downscale HR images to visually similar LR counterparts, which are subsequently reconstructed to high resolution. Early studies typically formulate this task as image super-resolution, employing bicubic interpolation for downscaling and learned super-resolution networks for upscaling~\citep{EDSR,RCAN,RealESRGAN}. Subsequent works seek to jointly optimize the downscaling and upscaling processes by replacing predefined downscaling operators with learnable models~\citep{TAD_TAU,CNN_CR,CAR}. To alleviate the ill-posed nature of upscaling, Xiao~et~al. presented invertible neural networks for image rescaling by modeling the lost information as a case-agnostic distribution~\citep{IRN}. Building upon this, Liang~et~al. enhanced high-frequency modeling by conditioning the flow-based framework on the LR image~\citep{HCFlow}. HfFlow~\citep{HfFlow} further incorporates reference LR images as prior knowledge to guide the downscaling process. Similarly, Bao~et~al. employed a plug-and-play tri-branch architecture that decomposes the low-frequency component into luminance and chrominance branches, while representing high-frequency components using all-zero mappings during upscaling~\citep{T-IRN}. 
Despite these advances, the reliance on relatively simple pixel-wise constraints limits their scalability, often leading to overly smooth reconstructions at large upscaling factors, e.g., $16\times$ or higher.

\subsection{Extreme Image Rescaling}
With the rapid advancement of ultra-high-definition imaging devices, extreme image rescaling, with scaling factors of $16\times$ or $32\times$, has become increasingly necessary. Zhong~et~al. proposed an extreme rescaling framework that incorporates StyleGAN priors to enable faithful reconstruction~\citep{GRAIN}; however, it is primarily limited to specific image domains, such as faces, churches, and cats. To generalize extreme rescaling to natural images with diverse content, Wei~et~al. exploited the high-quality visual embeddings encapsulated in the pretrained VQGAN~\citep{VQIR}. More recently, Wang~et~al. introduced an additional pixel-guidance module to enhance LR image generation~\citep{TADM}. 
Although the aforementioned methods attempt to address extreme image rescaling using generative priors, they still struggle to reconstruct high-fidelity and realistic images. 

By contrast, we propose a diffusion-based rescaling approach that leverages visual semantic priors from low-resolution thumbnails as strong guidance to steer diffusion reconstruction, leading to faithful rescaling.

\section{Proposed Method}\label{method}
\begin{figure*}[tbp]
\centering
\includegraphics[width=1.0\textwidth]{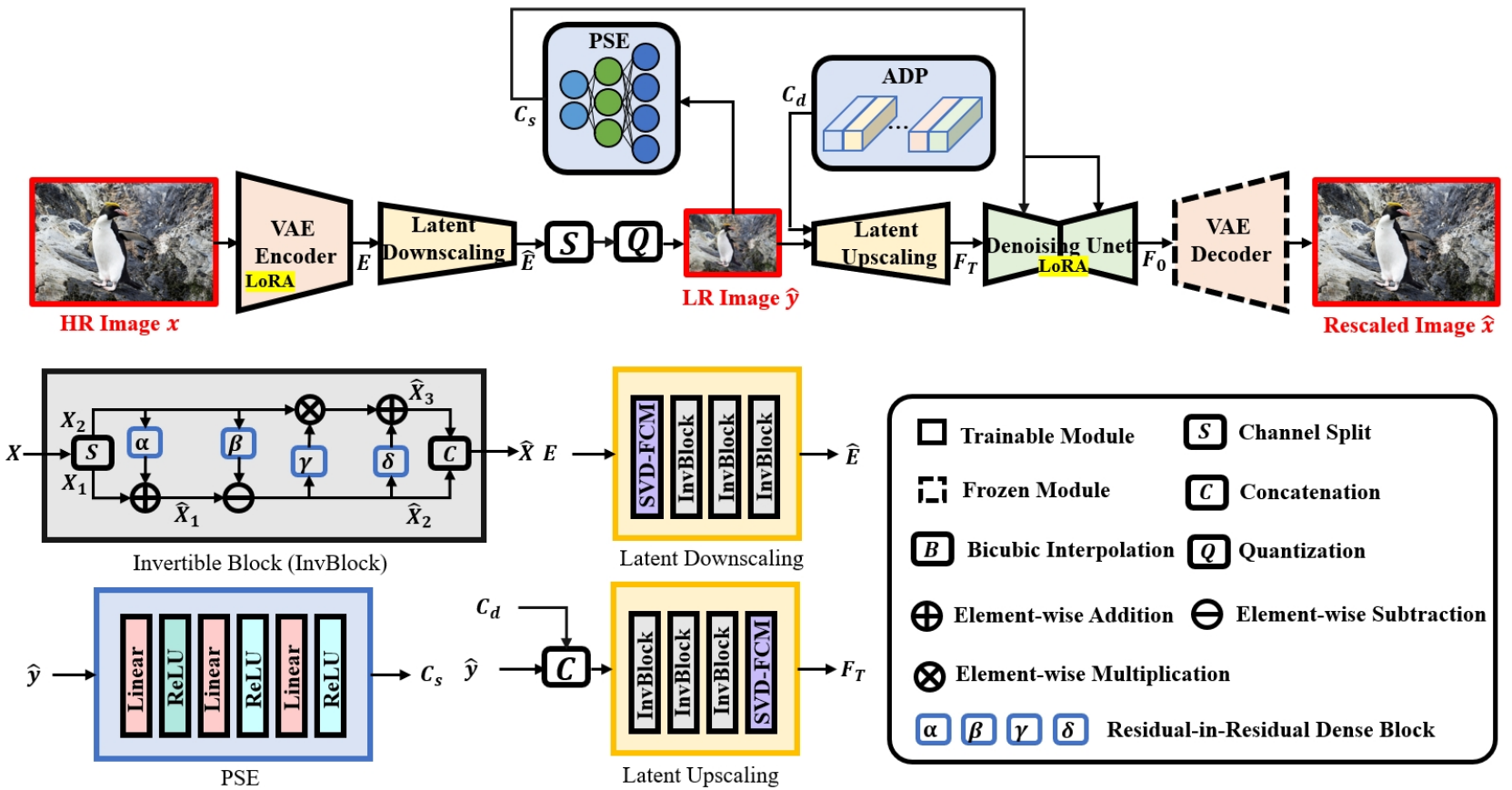}
\caption{Overview of FaithEIR. The VAE encoder maps the HR input into latent features. A latent rescaling module performs downscaling to produce the LR image and upscaling to recover the latent representation, where the ADP facilitates high-frequency compensation during upscaling. A pretrained diffusion model restores clean rescaled features conditioned on semantic priors derived from the PSE module. Finally, the frozen VAE decoder reconstructs the HR output.} 
\label{arch}
\end{figure*}
\subsection{Overview of FaithEIR}
Fig.~\ref{arch} shows the overall framework of FaithEIR. Given an input HR image $x$, we first employ a VAE encoder $\mathcal{E}_{\text{vae}}$ to map $x$ into a compact latent representation $\mathbf{E}$. The latent $\mathbf{E}$ is then processed by the latent downscaling module $\mathcal{E}_{\text{ld}}$, which comprises the proposed SVD-FCM module followed by three invertible blocks~\citep{SAIN}. 
Subsequently, a quantization operator $Q$ is applied to generate the LR image $\hat{y}$.
The overall downscaling process can be formulated as
\begin{equation}
\mathbf{E} = \mathcal{E}_{\text{vae}}(x), \quad
\hat{y} = Q\big(S(\mathcal{E}_{\text{ld}}(\mathbf{E}))\big),
\end{equation}
where $S(\cdot)$ denotes a channel-wise split operation, in which the first three features are used to generate the LR image, while the remaining features are discarded.

During the upscaling stage, the latent upscaling module $\mathcal{D}_{\text{lu}}$ takes the LR image $\hat{y}$ together with the proposed ADP $\mathbf{C_d}$ as input and produces a rescaled latent representation $\mathbf{F}_T$, which serves as a noisy latent for subsequent denoising. A one-step denoiser $\epsilon_{\text{SD}}$ then generates a clean latent $\mathbf{F}_0$ conditioned on semantic priors $\mathbf{C_s}$ extracted by the proposed PSE module. Finally, the reconstructed HR image $\hat{x}$ is produced using the frozen VAE decoder $\mathcal{D}_{\text{vae}}$.
The overall upscaling process can be expressed as
\begin{equation}
\label{up}
\mathbf{F}_T = \mathcal{D}_{\text{lu}}(\hat{y}, \mathbf{C_d}), \quad
\mathbf{F}_0 = \epsilon_{\text{SD}}(\mathbf{F}_T, \mathbf{C_s}), \quad
\hat{x} = \mathcal{D}_{\text{vae}}(\mathbf{F}_0).
\end{equation}

\subsection{Performing Rescaling in Latent Space}

\textbf{SVD-Initialized Free Channel Mixing.}
Downsampling is a fundamental component of image rescaling. Previous methods typically formulate downscaling as a decomposition into low- and high-frequency components using the Haar transform (HaarT)~\citep{IRN,T-IRN}. While mathematically elegant, this fixed transform is suboptimal for task-specific feature representation. To address this limitation, we propose an SVD-FCM module that performs adaptive, data-driven feature decomposition, enabling more effective preservation and separation of semantically meaningful information for extreme image rescaling.

The core component of the proposed SVD-FCM is a learnable non-orthogonal matrix that replaces the fixed HaarT kernel. During the forward (downsampling) pass, input features are first rearranged via a pixel-unshuffle operation (space-to-depth~\citep{ESPCN}). The rearranged features are then linearly transformed using the learnable matrix, which combines an orthogonal starting point via SVD initialization with unconstrained training. During the reverse (upsampling) pass, the transpose of the learned matrix is applied, followed by a pixel-shuffle operation to restore the original spatial resolution.

Specifically, the forward process of the proposed SVD-FCM can be formulated as:
\begin{equation}
\label{fp}
\mathbf{Y} = R\left(\mathbf{W}F\left(P(\mathbf{Z}, s)\right)\right),
\end{equation}
where $\mathbf{Z}$ and $\mathbf{Y}$ denote the input and output features, respectively; $P(\cdot, s)$ represents the pixel-unshuffle operation with scale factor $s$; $F(\cdot)$ and $R(\cdot)$ denote the flatten and reshape operations; and $\mathbf{W}$ is the learnable matrix. Note that $\mathbf{W}$ is initialized via SVD-based orthogonalization as
\begin{equation}
\mathbf{W} \leftarrow \mathbf{U}\mathbf{V}^\top, \quad \text{where } \mathbf{U}\boldsymbol{\Sigma}\mathbf{V}^\top = \mathrm{SVD}(\mathbf{W}).
\end{equation}

The backward pass of the proposed SVD-FCM corresponds to the inverse of Eq.~(\ref{fp}) and is formulated as:
\begin{equation}
\hat{\mathbf{Z}} = P^{-1}\left(R\left(\mathbf{W}^\top F(\mathbf{Y})\right), s\right),
\end{equation}
where $\hat{\mathbf{Z}}$ denotes the upsampled features, and $P^{-1}(\cdot, s)$ represents the pixel-shuffle operation with scale factor $s$.

%

\textbf{Adaptive Detail Prior.}
\begin{figure}[t]
\centering
\includegraphics[width=0.5\textwidth]{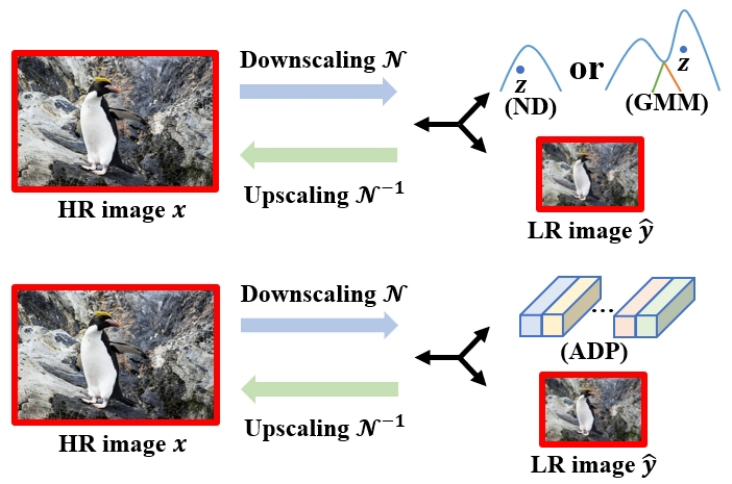}
\caption{Macro-architecture comparison between previous methods (top) and our method (bottom). Existing approaches model high-frequency components using ND or GMM. In contrast, our method introduces ADP to effectively compensate for high-frequency information loss.} 
\label{adp}
\end{figure}
Compensating for the loss of high-frequency information during downscaling remains a fundamental challenge in image rescaling. Traditional approaches often address this issue by probabilistically modeling the missing details, typically assuming either a normal distribution (ND)~\citep{IRN} or a learnable Gaussian Mixture Model (GMM)~\citep{SAIN}. As discussed in~\citep{T-IRN}, these strategies suffer from a key limitation that the inherent randomness of latent variables $z$ can lead to stochastic HR reconstructions that are misaligned with the deterministic objective of image rescaling. 

Based on these observations, we propose an effective deterministic approach, termed the adaptive detail prior (ADP), a learnable high-frequency dictionary that captures the empirical average of lost high-frequency components across the training dataset, thereby yielding a minimum mean-squared-error solution. From this perspective, the optimal compensation is not a stochastic distribution but a deterministic tensor. Specifically, the ADP is parameterized as a learnable tensor that stores characteristic high-frequency patterns. The effectiveness of the proposed ADP is demonstrated in Section~\ref{impact_adp}.

\subsection{Semantic Priors Coupled with Diffusion Model}
Leveraging semantic priors, such as text descriptions, semantic segmentation maps, and implicit priors, has been shown to enable more fine-grained and semantically consistent texture generation in perceptual image compression and super-resolution~\citep{DSSLIC,SeeSR,ICISP}. However, directly applying such semantic priors to extreme image rescaling remains challenging because severely downscaled LR images often lack sufficient semantic cues for accurate and reliable prior extraction. Moreover, relying on external captioning or segmentation models to generate semantic information substantially increases the computational complexity of the upscaling process, limiting the practicality of these approaches for rescaling tasks.

To this end, we introduce a lightweight pixel semantic embedder (PSE) to extract informative semantic cues $\mathbf{C_s}$ directly from the LR image $\hat{y}$. As shown in Fig.~\ref{arch}, the PSE consists of a small stack of fully connected layers with ReLU activations, resulting in a compact and computationally efficient design. Conditioned on the extracted semantic priors $\mathbf{C_s}$, we adopt a single-step diffusion denoising strategy that directly maps a noisy latent $\mathbf{F}_T$ to the clean latent $\mathbf{F}_0$, following the SD-Turbo paradigm~\citep{SD-Turbo}. The denoising process is formulated as

\begin{equation}
\mathbf{F}_0 = \frac{\mathbf{F}_T - \sqrt{1-\bar{\alpha}_T}\cdot\epsilon_{\text{SD}}(\mathbf{F}_T, \mathbf{C_s}, T)}{\sqrt{\bar{\alpha}_T}},
\end{equation}
where $T$ is fixed 
to emulate the maximum noise level, and $\bar{\alpha}_T$ denotes the cumulative noise coefficient of the DDPM scheduler~\citep{DDPM}. This design enables efficient single-step reconstruction while preserving the strong generative priors of diffusion models.

\subsection{Training Objectives}
To obtain visually plausible LR images $\hat{y}$, we introduce a LR guidance loss defined as
\begin{equation}
\mathcal{L}_{\text{lr}} = \lVert \hat{y} - \mathcal{B}(x) \rVert_2,
\end{equation}
where $\mathcal{B}(\cdot)$ denotes bicubic interpolation. To further regularize the PSE and encourage semantic consistency between the LR image $\hat{y}$ and its corresponding HR image $x$, we impose a semantic alignment loss
\begin{equation}
\mathcal{L}_{\text{sem}} = \lVert \mathbf{C_s} - \mathcal{E}_{\text{sem}}(x) \rVert_2,
\end{equation}
where $\mathcal{E}_{\text{sem}}$ is a pretrained frozen DINOv2 encoder~\citep{DINOv2} serving as a strong semantic feature extractor. In addition, we apply supervision to the reconstructed HR image $\hat{x}$ at both the pixel and feature levels, formulated as
\begin{equation}
\mathcal{L}_{\text{p}} = \lVert \hat{x} - x \rVert_2, \quad
\mathcal{L}_{\text{f}} = \lVert \phi(\hat{x}) - \phi(x) \rVert_2,
\end{equation}
where $\phi(\cdot)$ denotes features extracted from a pretrained VGG network.

The overall training objective is defined as a weighted sum of the above losses:
\begin{equation}
\label{loss_total}
\mathcal{L}_{\text{total}} = \lambda_1 \mathcal{L}_{\text{p}} + \lambda_2 \mathcal{L}_{\text{f}} + \lambda_3 \mathcal{L}_{\text{lr}} + \lambda_4 \mathcal{L}_{\text{sem}},
\end{equation}
where $\left\{\lambda_1, \lambda_2, \lambda_3, \lambda_4\right\}$ are scalar hyperparameters.

To narrow the distribution gap between reconstructed HR image $\hat{x}$ and original HR image $x$, we fine-tune FaithEIR
using adversarial training by introducing a discriminator. 
The generator loss is defined as:
\begin{equation}
    \mathcal{L}_{\text{gen}} = \mathcal{L}_{\text{total}} + \lambda_5 \mathcal{L}_{\text{adv}},
\end{equation}
where $\lambda_5$ is set to 0.05 to balance the adversarial loss $\mathcal{L}_{\text{adv}}$. The adversarial loss is expressed as:
\begin{equation}
    \mathcal{L}_{\text{adv}}=-\mathbb{E}(\mathcal{H}(\hat{x})),
\end{equation}
where $\mathcal{H}$ denotes the discriminator. To regularize the discriminator, the discriminator loss is formulated as:
\begin{equation}
    \mathcal{L}_{\text{dis}} = \mathbb{E}(log\mathcal{H}(x))+\mathbb{E}(log(1-\mathcal{H}(\hat{x}))).
\end{equation}
\section{Experiments}\label{experiments}
\subsection{Experimental Setup}
\begin{table*}[t]
\centering
\caption{Quantitative comparisons of HR reconstruction performance for different methods at 16$\times$ and 32$\times$ scaling factors (separated by ``$/$") on benchmark datasets. The symbols $\uparrow$ and $\downarrow$ indicate that higher and lower values correspond to better performance, respectively. The best and second-best results are marked in bold and underline, respectively.}
\setlength{\tabcolsep}{2pt}
\begin{adjustbox}{max width=\textwidth}
\begin{tabular}{ccccccccc}
\toprule
Dataset & Method & LPIPS $\downarrow$ & DISTS $\downarrow$ & FID $\downarrow$ & KID $\downarrow$ & PSNR $\uparrow$ & SSIM $\uparrow$ & MUSIQ$\uparrow$ \\
\midrule
\multirow{9}{*}{DIV2K-val} & SeD & 0.4710 / 0.5725 & 0.2160 / 0.3574 & 65.46 / 127.27 & 0.0183 / 0.0568 & 20.92 / 18.56  & 0.5412 / 0.4599 & 63.80 / 55.75 \\
    & S3Diff & 0.4287 / 0.5430 & 0.1577 / \underline{0.2264} & 27.67 / 58.96 & 0.0029 / \underline{0.0146} & 18.87 / 16.20 & 0.4909 / 0.3961 & \textbf{69.55} / 65.47 \\
    & DiffBIR & 0.4835 / 0.5825 & 0.2468 / 0.3189 & 68.49 / 106.76  & 0.0180 / 0.0341 & 21.10 / 18.51 & 0.5185 / 0.3880 & 66.48 / \underline{67.18}\\ 
    & IRN  & 0.5000 / 0.6162 & 0.3014 / 0.4139 & 130.04 / 168.52 & 0.0636 / 0.0979  & \underline{24.78} / \underline{21.19}   & \underline{0.6856} / \underline{0.5774} & 47.41 / 27.02\\
    & T-IRN   & 0.4984 / 0.5895 & 0.2978 / 0.3945 & 126.10 / 160.83 & 0.0605 / 0.0900  & \textbf{24.80} / \textbf{21.90} & \textbf{0.6882} / \textbf{0.6020} & 47.37 / 36.45\\
    & VQIR     & \underline{0.3018} / \underline{0.4583} & \textbf{0.0877} / 0.2436  & \underline{16.09} / \underline{81.06} & \textbf{0.0013} / 0.0285 & 22.19 / 20.20  & 0.6201 / 0.5414 & 66.10 / 58.29\\
    & TADM      & 0.3602 / 0.4765 & 0.1808 / 0.2721 & 49.05 / 110.41 & 0.0148 / 0.0577 & 22.08 / 20.35  & 0.6164 / 0.5338 & 62.44 / 57.88\\
    & FaithEIR & \textbf{0.2900} / \textbf{0.3751} & \underline{0.0902} / \textbf{0.1421} & \textbf{13.42} / \textbf{21.60} &  \textbf{0.0013} / \textbf{0.0018} & 21.97 / 20.29  & 0.6205 / 0.5439  & \underline{68.50} / \textbf{67.75}\\ 
\midrule
\multirow{9}{*}{LSDIR-val} & SeD & 0.4888 / 0.5900  & 0.2279 / 0.3588  & 57.44 / 129.22  & 0.0162 / 0.0596 & 18.40 / 16.70   & 0.4446 / 0.3765 & 68.12 / 59.80\\
    & S3Diff & 0.4350 / 0.5526 & 0.1630 / \underline{0.2304} & 22.44 / \underline{46.53} & 0.0026 / \underline{0.0117} & 16.96 / 14.82 & 0.4112 / 0.3307 & 71.99 / 68.48\\
    & DiffBIR & 0.4868 / 0.5807 & 0.2277 / 0.2973 & 53.08 / 81.18 & 0.0157 / 0.0275 & 18.39 / 16.80 & 0.4347 / 0.3397 & 70.07 / \underline{69.50}\\ 
    & IRN  & 0.5346 / 0.6645 & 0.3253 / 0.4360 & 142.12 / 191.95 & 0.0798 / 0.1268 & \textbf{21.41} / \underline{18.66}   & \underline{0.5863} / \underline{0.4654}  & 54.42 / 26.93\\
    & T-IRN   & 0.5329 / 0.6324 & 0.3221 / 0.4168 & 139.02 / 177.60 & 0.0779 / 0.1118  & \textbf{21.41} / \textbf{19.22} & \textbf{0.5895} / \textbf{0.4937} & 54.11 / 40.74\\
    & VQIR    & \underline{0.3231} / \underline{0.4865} & \underline{0.1128} / 0.2866 & \underline{15.79} / 86.53 & \underline{0.0018} / 0.0343  & 19.09 / 17.87  & 0.5231 / 0.4276 & \underline{72.02} / 60.82\\
    & TADM     & 0.3955 / 0.5158 & 0.2111 / 0.3044 & 54.30 / 120.65 & 0.0212 / 0.0723 & 19.09 / 17.90 & 0.5046 / 0.4215 & 68.25 / 62.28 \\
    & FaithEIR & \textbf{0.2874} / \textbf{0.3869} & \textbf{0.0933} / \textbf{0.1664} & \textbf{10.64} / \textbf{19.41}  & \textbf{0.0008} / \textbf{0.0021} & 19.68 / 18.06 & 0.5431 / 0.4522 & \textbf{73.11} / \textbf{72.05}\\
\midrule
\multirow{9}{*}{CLIC2020} & SeD & 0.4478 / 0.5450 & 0.1971 / 0.3181 & 38.06 / 90.32 & 0.0125 / 0.0416 & 23.54 / 20.87 & 0.6578 / 0.5985 & 60.30 / 52.97  \\ 
    & S3Diff & 0.4210 / 0.5272 & 0.1591 / 0.2227 & 13.27 / 35.41 & 0.0032 / \underline{0.0132} & 20.94 / 18.01 & 0.5949 / 0.5164 & \textbf{66.99} / \underline{63.56}\\
    & DiffBIR & 0.4689 / 0.5627 & 0.2267 / 0.3078 & 39.51 / 71.93 & 0.0136 / 0.0290 & 23.38 / 21.00 & 0.6428 / 0.5173 & 61.28 / \textbf{63.75}\\ 
    & IRN  & 0.4864 / 0.5808 & 0.2796 / 0.3913 & 86.91 / 127.48 & 0.0429 / 0.0760 & \underline{27.60} / \underline{23.72}  & \underline{0.7659} / \underline{0.6895} & 47.24 / 26.26\\
    & T-IRN    & 0.4846 / 0.5586 & 0.2752 / 0.3709 & 83.37 / 109.77 & 0.0411 / 0.0598 & \textbf{27.62} / \textbf{24.69} & \textbf{0.7670} / \textbf{0.7095} & 46.97 / 37.27\\
    & VQIR     & \underline{0.2804} / \underline{0.4245} & \underline{0.0828} / 0.2241 & \underline{6.87} / \underline{46.31} & \underline{0.0014} / 0.0171 & 24.84 / 22.56 & 0.7117 / 0.6550 & 60.55 / 52.24\\
    & TADM     & 0.3320 / 0.4447 & 0.1615 / 0.2460 & 28.48 / 74.15 & 0.0106 / 0.0395 & 24.52 / 22.61  & 0.7051 / 0.6445 & 56.02 / 52.53\\
    & FaithEIR & \textbf{0.2706} / \textbf{0.3476} & \textbf{0.0806} / \textbf{0.1274} & \textbf{5.33} / \textbf{8.93} & \textbf{0.0011} / \textbf{0.0015} & 24.13 / 22.74 & 0.7016 / 0.6532 & \underline{63.73} / 63.19\\
\bottomrule
\end{tabular}
\end{adjustbox}
\label{quan_com}
\end{table*}
\subsubsection{Dataset} 
We train the proposed FaithEIR on the \textit{LSDIR} dataset~\citep{LSDIR}, which contains 84,995 images covering diverse scenes. For evaluation, we conduct experiments on three high-resolution benchmark datasets: \textit{DIV2K-val}~\citep{DIV2K}, \textit{LSDIR-val}~\citep{LSDIR}, and \textit{CLIC2020}\\~\citep{CLIC2020}.

\subsubsection{Evaluation metrics}
To quantitatively evaluate the performance of image rescaling methods, we adopt \textit{LPIPS}~\citep{LPIPS} and \textit{DISTS}~\citep{DISTS} as full-reference perceptual quality metrics. The no-reference perceptual quality metric \textit{MUSIQ}~\citep{MUSIQ_ICCV2021} is also included as a supplementary metric. In addition, we employ \textit{FID}~\citep{FID} and \textit{KID}~\citep{KID} to measure the distribution discrepancy between the reconstructed and original images. Following the evaluation protocol in~\citep{VQIR,ICISP}, we compute FID and KID on 256$\times$256 image patches to ensure reliable assessment. For completeness, we also report standard fidelity measures, including \textit{PSNR} and \textit{SSIM}~\citep{SSIM}.

\subsubsection{Implementation details}
We implement our method using PyTorch and train it on a machine equipped with two NVIDIA GeForce RTX 4090 GPUs. During training, images are randomly cropped to 512$\times$512, and the batch size is set to 8. We adopt the AdamW~\citep{Adam} optimizer with default hyperparameter settings. The model is trained for over 300K iterations with an initial learning rate of $1e^{-4}$, which is halved every 100K iterations. Our rescaling framework is based on SD-Turbo~\citep{SD-Turbo} and is fine-tuned using LoRA, with rank parameters set to 16 for the VAE encoder and 32 for the denoising UNet. The hyperparameters $\lambda_1, \lambda_2, \lambda_3, \lambda_4$ in Eq.(\ref{loss_total}) are set to 2.0, 5.0, 3.0, and 3.0. For adversarial training, we use a pretrained DINO~\citep{DINO} as the backbone of the discriminator, and introduce multiple independent classifiers corresponding to different feature levels to distinguish images more effectively. The learning rates of $1e^{-6}$ and $1e^{-4}$ to the FaithEIR (generator) and discriminator, respectively. Their network parameters are updated for around 22K iterations using the AdamW~\citep{Adam} optimizer.

\subsection{Comparisons with State-of-the-art Methods}
We compare the proposed approach with state-of-the-art super-resolution and rescaling methods, including GAN-based SR approaches (e.g., SeD~\citep{SeD}), diffusion-based SR methods (e.g., S3Diff~\citep{S3Diff} and DiffBIR~\citep{DiffBIR}), rescaling methods using invertible neural networks (e.g., IRN~\citep{IRN} and T-IRN~\citep{T-IRN}), and prior-based rescaling methods (e.g., VQIR~\citep{VQIR} and TADM~\citep{TADM}).

Table~\ref{quan_com} summarizes the quantitative evaluation results. 
Our method consistently outperforms competing approaches in terms of perception-oriented metrics such as LPIPS, DISTS, FID, and KID. 
Although T-IRN achieves the best PSNR and SSIM scores, its perceptual scores are significantly poorer than those of our method.

\begin{figure*}[t]
\centering
\footnotesize
\setlength{\tabcolsep}{1pt} 
\begin{adjustbox}{max width=\textwidth}
\begin{tabular}{@{}c c c c c@{}}
\includegraphics[width=0.2\textwidth]{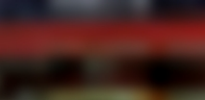} &
\includegraphics[width=0.2\textwidth]{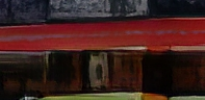} &
\includegraphics[width=0.2\textwidth]{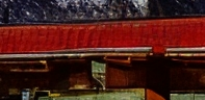} &
\includegraphics[width=0.2\textwidth]{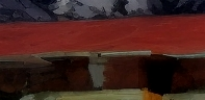} &
\includegraphics[width=0.2\textwidth]{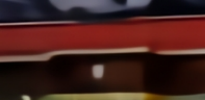} \\
(a) Bicubic & (b) SeD & (c) S3Diff & (d) DiffBIR & (e) IRN \\
(18.68 / 0.6346) & (17.94 / 0.4401) & (16.57 / 0.4314) & (18.55 / 0.4693) & (\textbf{20.73} / 0.5031) \\
\includegraphics[width=0.2\textwidth]{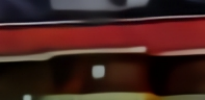} &
\includegraphics[width=0.2\textwidth]{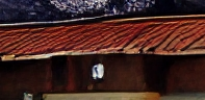} &
\includegraphics[width=0.2\textwidth]{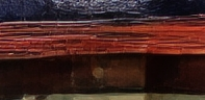} &
\includegraphics[width=0.2\textwidth]{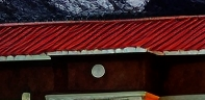} &
\includegraphics[width=0.2\textwidth]{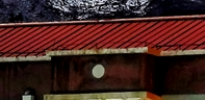} \\
(f) T-IRN & (g) VQIR & (h) TADM & (i) FaithEIR & (j) HR patch \\
(20.70 / 0.5038) & (18.94 / 0.3024) & (18.50 / 0.3644) & (19.05 / \textbf{0.2704}) & (PSNR / LPIPS) \\
\end{tabular}
\end{adjustbox}
\caption{Visual comparisons with state-of-the-art methods on the LSDIR-val dataset (\textbf{16}$\times$ HR reconstruction). Compared to other methods, FaithEIR generates more realistic results with faithful structures.}
\label{vis_com_16}
\end{figure*}

\begin{figure*}[t]
\centering
\footnotesize
\setlength{\tabcolsep}{1pt} 
\begin{adjustbox}{max width=\textwidth}
\begin{tabular}{@{}c c c c c@{}}
\includegraphics[width=0.2\textwidth]{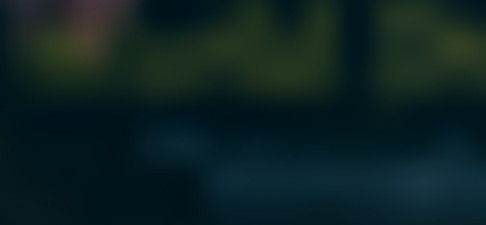} &
\includegraphics[width=0.2\textwidth]{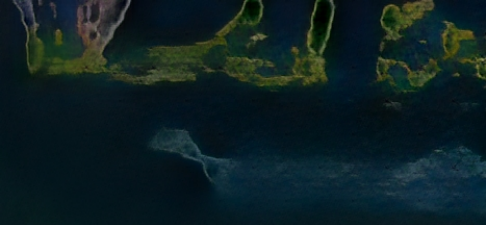} &
\includegraphics[width=0.2\textwidth]{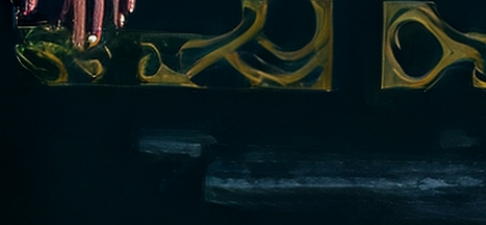} &
\includegraphics[width=0.2\textwidth]{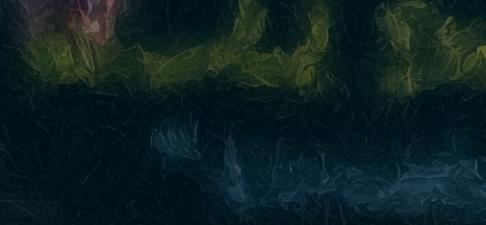} &
\includegraphics[width=0.2\textwidth]{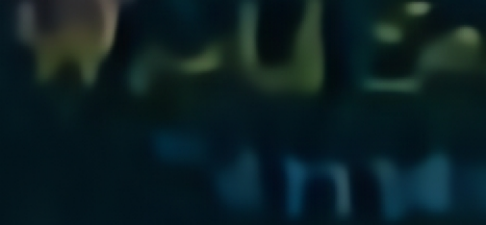} \\
(a) Bicubic & (b) SeD & (c) S3Diff & (d) DiffBIR & (e) IRN \\
(20.89 / 0.4951) & (19.41 / 0.5139) & (17.19 / 0.5250)  & (20.51 / 0.6439) & (22.48 / 0.4436) \\
\includegraphics[width=0.2\textwidth]{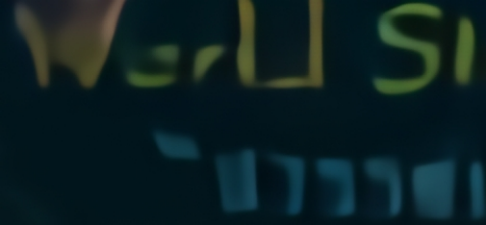} &
\includegraphics[width=0.2\textwidth]{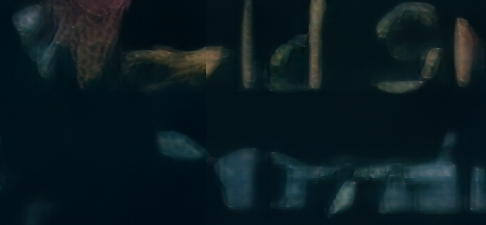} &
\includegraphics[width=0.2\textwidth]{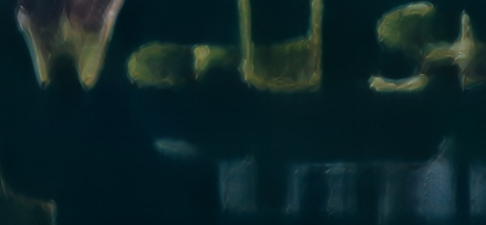} &
\includegraphics[width=0.2\textwidth]{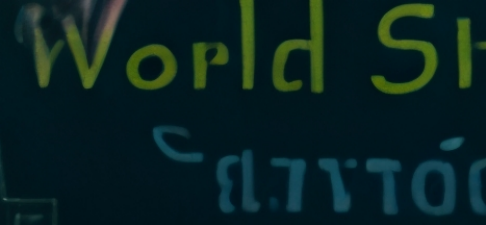} &
\includegraphics[width=0.2\textwidth]{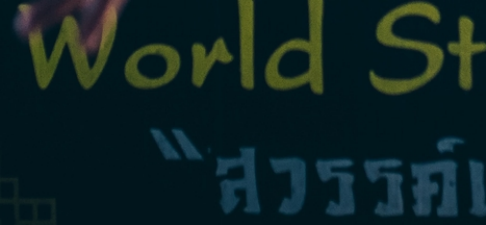} \\
(f) T-IRN & (g) VQIR & (h) TADM & (i) FaithEIR & (j) HR patch \\
(\textbf{23.30} / 0.3937) & (21.53 / 0.3875) & (22.08 / 0.4039) & (21.92 / \textbf{0.3049}) & (PSNR / LPIPS) \\
\end{tabular}
\end{adjustbox}
\caption{Visual comparisons with state-of-the-art methods on the CLIC2020 dataset (\textbf{32}$\times$ HR reconstruction). Compared to other methods, FaithEIR generates better results, where the letters are recognizable.}
\label{vis_com_32}
\end{figure*}

Fig.~\ref{vis_com_16} and \ref{vis_com_32} present visual comparisons on benchmark datasets. The results produced by SR–based methods appear overly smoothed (see (b)-(d) of Fig.~\ref{vis_com_16} and \ref{vis_com_32}). Previous rescaling methods based on invertible neural networks and generative priors either yield incorrect structures or introduce noticeable artifacts. In contrast, our method produces more realistic reconstructions with faithful structures and finer details; for example, the structure of the house is well preserved in Fig.~\ref{vis_com_16}(i), and the letters are clearly recognizable in Fig.~\ref{vis_com_32}(i).
\subsection{Discussions and Analysis}
%
\begin{table}
\centering
\caption{Effectiveness of the proposed SVD-FCM and ADP.}
\begin{tabular}{cccc}
\toprule
Exp. & Method  & LPIPS $\downarrow$ & DISTS $\downarrow$\\
\midrule
E1 & w/ HaarT+ND~\citep{IRN} & 0.6014   & 0.2827     \\
E2 & w/ HaarT+GMM~\citep{SAIN} & 0.4507  & 0.2065       \\
E3 & w/ HaarT+ADP & 0.6047 & 0.2901 \\
E4 & w/ Cayley+ADP & 0.3324 & 0.1267 \\ 
E5 & w/ SVD-FCM+ADP (Ours) & \textbf{0.3230} & \textbf{0.1218} \\
\bottomrule
\end{tabular}
\label{ablation_exp}
\end{table}
\begin{figure}[!t]
\footnotesize
\centering
    \begin{tabular}{cccc}
            \includegraphics[width=0.19\linewidth,height=0.19\linewidth]{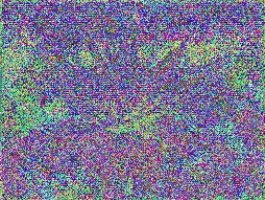}
            \includegraphics[width=0.19\linewidth,height=0.19\linewidth]{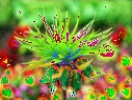}
            \includegraphics[width=0.19\linewidth,height=0.19\linewidth]{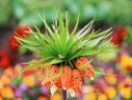}
            \includegraphics[width=0.19\linewidth,height=0.19\linewidth]{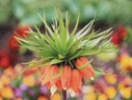} 
            \includegraphics[width=0.19\linewidth,height=0.19\linewidth]{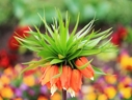} \\
            \makebox[0.19\linewidth]{(a) HaarT+ND}
            \makebox[0.19\linewidth]{(b) HaarT+GMM} 
            \makebox[0.19\linewidth]{(c) HaarT+ADP} 
            \makebox[0.19\linewidth]{(d) SVD-FCM+ADP} 
            \makebox[0.19\linewidth]{(e) Reference LR} \\
            \includegraphics[width=0.19\linewidth,height=0.19\linewidth]{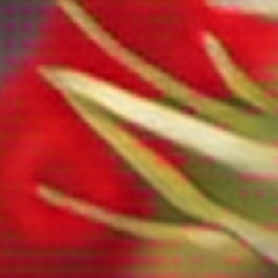}
            \includegraphics[width=0.19\linewidth,height=0.19\linewidth]{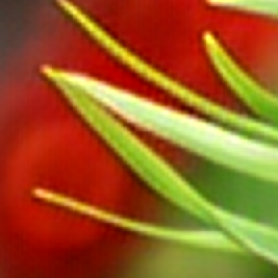} 
            \includegraphics[width=0.19\linewidth,height=0.19\linewidth]{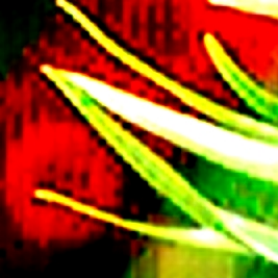}
            \includegraphics[width=0.19\linewidth,height=0.19\linewidth]{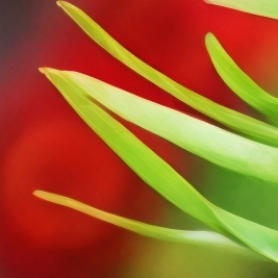} 
            \includegraphics[width=0.19\linewidth,height=0.19\linewidth]{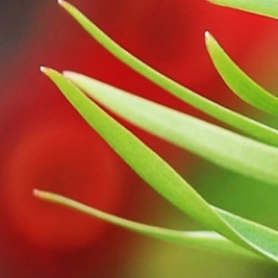} \\
            \makebox[0.19\linewidth]{(f) HaarT+ND}
            \makebox[0.19\linewidth]{(g) HaarT+GMM} 
            \makebox[0.19\linewidth]{(h) HaarT+ADP} 
            \makebox[0.19\linewidth]{(i) SVD-FCM+ADP} 
            \makebox[0.19\linewidth]{(j) HR patch} \\
    \end{tabular}
\caption{Effectiveness of SVD-FCM and ADP. (a)-(d) and (f)-(i) are generated LR images and reconstruction HR results, respectively.}
\label{vis_com_lrt}
\end{figure}
\subsubsection{Effectiveness of SVD-FCM}
The proposed SVD-FCM is designed to adaptively perform rescaling in the latent space. To validate its effectiveness, we compare it with the Haar transform, widely used in existing rescaling methods. As shown in Table~\ref{ablation_exp}, SVD-FCM outperforms HaarT, achieving lower LPIPS and DISTS scores (E3 vs. E5). Moreover, Fig.~\ref{vis_com_lrt}(c) and (d) illustrate that SVD-FCM produces more visually pleasing LR reconstructions. Fig.~\ref{vis_com_lrt}(h) and (i) further demonstrate that SVD-FCM contributes more effectively than HaarT to high-quality HR reconstruction.

In addition, we apply a Cayley transform to the proposed SVD-FCM to make the learnable matrix orthogonal. As shown in Table~\ref{ablation_exp} (E4 vs. E5), using an orthogonality constraint does not improve the reconstruction performance. In contrast, despite losing exact invertibility, the proposed SVD-FCM achieves lower LPIPS and DISTS scores. We attribute this to the loss of scaling freedom: orthogonal matrices cannot compress redundant channels during downsampling or amplify informative channels during upsampling.

\subsubsection{Effect of ADP}\label{impact_adp}
Unlike previous methods that model missing high-frequency components using ND or GMM, we propose ADP to compensate for information loss. As shown in Table~\ref{ablation_exp}, ADP fails to improve HR reconstruction when the HaarT is used for rescaling, resulting in higher LPIPS and DISTS scores. In contrast, when rescaling is performed with the proposed SVD-FCM, ADP yields significant performance gains (E3 vs. E5). The reason behind this phenomenon the fixed HaarT restricts the ability of ADP to adaptively learn meaningful structures, while SVD-FCM produces a data-adaptive representation space, enabling ADP to better learn dictionary atoms and capture structural patterns, thereby improving reconstruction quality.
The visual results in Fig.~\ref{vis_com_lrt} further support this observation: ADP improves the quality of LR reconstructions, and its effectiveness for HR reconstruction becomes evident only when coupled with SVD-FCM.

\begin{figure}[!t]
\footnotesize
\centering
    \begin{tabular}{cc}
            \includegraphics[width=0.48\linewidth,height=0.44\linewidth]{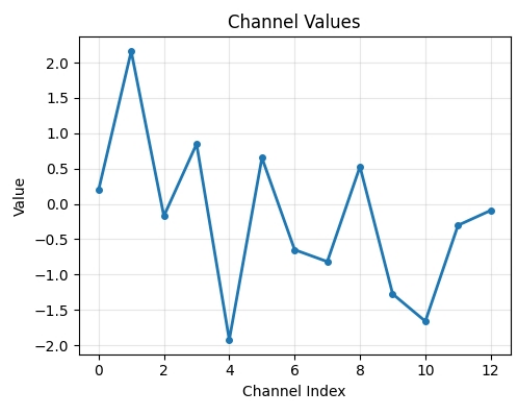}
            \includegraphics[width=0.48\linewidth,height=0.44\linewidth]{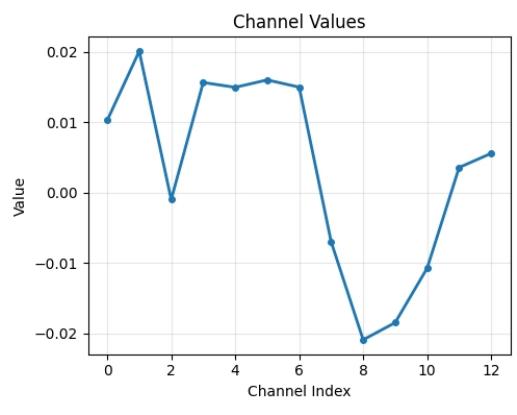} \\
            \makebox[0.48\linewidth]{(a) Random ADP} 
            \makebox[0.48\linewidth]{(b) Learnable ADP} \\
    \end{tabular}
\caption{Visualization of different ADP values.}
\label{vis_com_adp}
\end{figure}

We further investigate the information captured by ADP. As illustrated in Fig.~\ref{vis_com_adp}, ADP learns the average missing details, which approximately follow a uniform distribution within the range $\left(-0.02, 0.02\right)$. Table~\ref{ablation_exp_adp} further demonstrates that the learnable ADP achieves superior HR reconstruction performance compared with the variant initialized with random values. 

\begin{table}
\centering
\caption{Impact of different ADP values on HR reconstruction.}
\begin{tabular}{ccccc}
\toprule
Method &  PSNR $\uparrow$ & SSIM $\uparrow$ & LPIPS $\downarrow$ & DISTS $\downarrow$\\
\midrule
Random ADP & 11.99 & 0.1958 & 0.7811 & 0.5750 \\
Learnable ADP & \textbf{22.24} & \textbf{0.6021} & \textbf{0.3230}   & \textbf{0.1218}     \\
\bottomrule
\end{tabular}
\label{ablation_exp_adp}
\end{table}

\subsubsection{Impact of SP}
\begin{figure}[!t]
\footnotesize
\centering
    \begin{tabular}{cccc}
            \includegraphics[width=0.235\linewidth,height=0.235\linewidth]{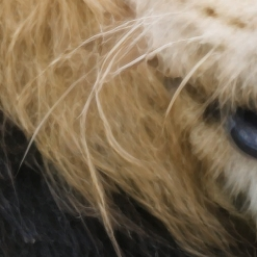}
            \includegraphics[width=0.235\linewidth,height=0.235\linewidth]{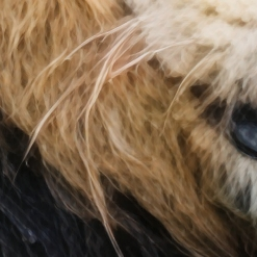}
            \includegraphics[width=0.235\linewidth,height=0.235\linewidth]{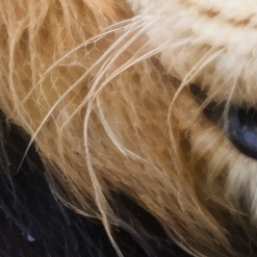}
            \includegraphics[width=0.235\linewidth,height=0.235\linewidth]{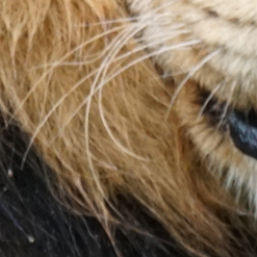} \\
            \makebox[0.235\linewidth]{(a) w/o SP}
            \makebox[0.235\linewidth]{(b) w/ LTE} 
            \makebox[0.235\linewidth]{(c) w/ SP} 
            \makebox[0.235\linewidth]{(d) HR patch} \\
    \end{tabular}
\caption{Impact of semantic priors (SP). Using effective semantic priors leads to reconstructions with more faithful details.}
\label{vis_com_sp}
\end{figure}

We compare our model with two variants: one without SP and another that employs learnable text embeddings (LTE) as SP~\cite{OSCAR}, to evaluate the impact of the proposed SP. As shown in Table~\ref{ablation_sp_exp}, the variant using learnable text embeddings fails to improve reconstruction performance; in particular, it produces a higher LPIPS score than the variant without SP (E1 vs. E2). In contrast, our model achieves the best overall performance, attaining the lowest LPIPS and DISTS scores (see E5). Fig.~\ref{vis_com_sp} further demonstrates that incorporating SP leads to superior reconstruction quality, with fine details such as the tiger’s whiskers being well preserved.   
\begin{table}
\centering
\caption{Effectiveness of the proposed visual semantic priors. LTE denotes the \underline{l}earnable \underline{t}ext \underline{e}mbeddings.}
\begin{tabular}{cccc}
\toprule
Exp. & Method  & LPIPS $\downarrow$ & DISTS $\downarrow$\\
\midrule
E1 & w/o SP & 0.3045 & 0.1078 \\
E2 & w/ LTE & 0.3230 & 0.1218 \\
E3 & w/ SP (layer4) & 0.3000 & 0.1024 \\
E4 & w/ SP (layer16) & 0.3058 & 0.1016 \\
E5 & w/ SP (layer23) & \textbf{0.2948} & \textbf{0.0993} \\
\bottomrule
\end{tabular}
\label{ablation_sp_exp}
\end{table}

\subsubsection{The semantics of different layers} 
It is worth noting that semantic representations extracted from different layers of the DINOv2 encoder~\citep{DINOv2} capture varying levels of abstraction. This naturally raises the question of which layer provides the most suitable semantic features for guiding the learning of the PSE module. To investigate this, we conduct ablation experiments summarized in Table~\ref{ablation_sp_exp}. We observe that semantic features from deeper layers yield greater performance gains, as they provide richer semantic blueprints that are more compatible with diffusion models for guiding the generation process (E3 vs. E4 vs. E5).

\subsubsection{Model complexity comparison}
\begin{table}
\centering
\caption{Model complexity comparisons on the DIV2K-val dataset (first 10 images) with a $16\times$ scale factor.}
\begin{tabular}{cccc}
\toprule
Method  & Downscaling (s) & Upscaling (s) & Parameters (M) \\
\midrule
IRN & 0.4032 & 0.4960 & 34.2     \\
T-IRN & 0.2173 & \textbf{0.4079} & \textbf{9.95}      \\
VQIR & 0.4867 & 2.2671 & 84.93 \\
TADM & 0.5561 & 1.6837 & 1,354.52 \\
FaithEIR & \textbf{0.1609} & 5.7912 & 985.03 \\
\bottomrule
\end{tabular}
\label{ITC}
\end{table}
We further evaluate the model complexity of the proposed method against state-of-the-art approaches, focusing on inference time and network parameters. As shown in Table~\ref{ITC}, our method achieves the fastest downscaling speed, while requiring relatively more computational time for upscaling. In addition, our model contains fewer network parameters than the diffusion-based competitor TADM, demonstrating a more efficient design.

\subsubsection{Comparison with JPEG compression}
JPEG compression is a widely adopted standard that regulates reconstruction quality through a quality factor (QF) ranging from 1 to 100, with higher QF providing better visual fidelity at the expense of a lower compression ratio. To demonstrate the storage efficiency of FaithEIR, we compare it with JPEG compression with respect to storage consumption and compression ratio. 
As shown in Table~\ref{ICC}, FaithEIR outperforms JPEG compression (QF = 1) by achieving higher reconstruction quality, reflected by a lower LPIPS score, while simultaneously reducing storage consumption by approximately 119,409 KB and improving the compression ratio by a factor of 192.57.
\begin{table}
\centering
\caption{Quantitative comparisons with JPEG compression on the DIV2K-val dataset. QF and CR denote the quality factor used in JPEG compression and compression ratio, respectively.}
\begin{tabular}{cccc}
\toprule
Method  & Storage (KB) & CR & LPIPS $\downarrow$ \\
\midrule
w/o JPEG & 477,418.32 & - & - \\
JPEG (QF=1) & \underline{120,037.44} & \underline{3.98} & 0.6500 \\
JPEG (QF=28) & 378,306.82 & 1.26 & 0.3004 \\
JPEG (QF=30) & 380,044.97 & 1.26 & \textbf{0.2912} \\
FaithEIR (16$\times$) & \textbf{628.16} & \textbf{196.55} &  \underline{0.2951} \\
\bottomrule
\end{tabular}
\label{ICC}
\end{table}
\begin{figure}[htbp]
\footnotesize
\centering
    \begin{tabular}{ccc}
            \includegraphics[width=0.31\linewidth,height=0.31\linewidth]{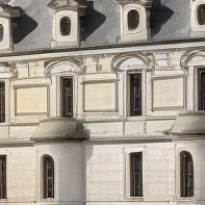}
            \includegraphics[width=0.31\linewidth,height=0.31\linewidth]{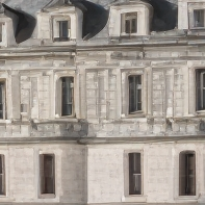} 
            \includegraphics[width=0.31\linewidth,height=0.31\linewidth]{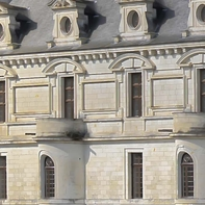} \\
            \makebox[0.31\linewidth]{(a) FaithEIR (16$\times$)} 
            \makebox[0.31\linewidth]{(b) FaithEIR (32$\times$)} 
            \makebox[0.31\linewidth]{(c) HR patch} \\
    \end{tabular}
\caption{Failure case: scenarios where our method produces semantically plausible but not strictly accurate details.}
\label{vis_com_lim}
\end{figure}
\subsubsection{Failure Case}
We present a failure case in Fig.~\ref{vis_com_lim}. In regions with highly ambiguous textures (e.g., building structures), the model may produce perceptually plausible yet not strictly accurate details, reflecting inherent uncertainty rather than semantic errors.
\section{Conclusion}
\label{conclusion}
In this paper, we propose an extreme image rescaling approach, termed FaithEIR, which performs rescaling in the latent space. We develop an SVD-initialized free channel mixing and an adaptive detail prior to enable effective latent feature rescaling and compensation for information loss, respectively. These two components collaborate to facilitate high-quality reconstruction at both the downscaling and upscaling stages. To effectively guide the diffusion model, we further design a pixel semantic embedder that extracts semantic conditions from LR images. Extensive quantitative and qualitative experiments demonstrate that the proposed approach performs favorably against state-of-the-art methods.
\Hao{Moreover, its downscaling speed surpasses that of previous comparison rescaling methods. However, it incurs high upscaling complexity due to the inherent computational overhead of generative models, rendering it unsuitable for deployment on resource-constrained receiver devices. Future work will explore model compression techniques, such as pruning, to mitigate this issue.}
\section{Acknowledgements}
This work are supported by the National Natural Science Foundation of China (NSFC62376208, U24A20265), Key Core Technology Research Project in Shaanxi Province (2024SF-GJHX-27) and Qin Chuangyuan General Window ``Four Chain'' Integration Project (2024PT-ZCK-66).
\bibliographystyle{elsarticle-num}
\bibliography{ref}




\end{document}